\documentclass{article}
\usepackage{iclr2026_conference,times}

\usepackage{amsmath,amsfonts,bm}

\def\eqref#1{equation~\ref{#1}}

\def\1{\bm{1}}

\DeclareMathAlphabet{\mathsfit}{\encodingdefault}{\sfdefault}{m}{sl}
\SetMathAlphabet{\mathsfit}{bold}{\encodingdefault}{\sfdefault}{bx}{n}

\usepackage{url}
\usepackage{amssymb}
\usepackage{booktabs}
\usepackage{multirow}
\usepackage{xcolor}

\usepackage{enumitem}
\usepackage{graphicx}
\usepackage{float}
\usepackage{caption}
\usepackage{subcaption}
\usepackage{pgfplots}
\pgfplotsset{compat=1.18}
\usepackage[colorlinks=true,citecolor=blue,linkcolor=blue,urlcolor=blue]{hyperref}
\usepackage{microtype}

\newcommand{\passone}{\text{pass@1}}

\newcommand{\passkk}{\text{pass@}k}
\newcommand{\bench}{\textsc{Riemann-Bench}}

\title{Riemann-Bench: A Benchmark for Moonshot Mathematics}
\iclrfinalcopy
\author{Suhaas Garre\thanks{Correspondence to: \texttt{suhaas@surgehq.ai}}, Erik Knutsen, Sushant Mehta, Edwin Chen  \\   
Surge AI \\   
}
\begin{document}

\maketitle

\begin{abstract}
Recent AI systems have achieved gold-medal-level performance on the International Mathematical Olympiad, demonstrating remarkable proficiency at competition-style problem solving. However, competition mathematics represents only a narrow slice of mathematical reasoning: problems are drawn from limited domains, require minimal advanced machinery, and can often reward insightful tricks over deep theoretical knowledge. We introduce \bench{}, a private benchmark of 25 expert-curated problems designed to evaluate AI systems on research-level mathematics that goes far beyond the olympiad frontier. Problems are authored by Ivy League mathematics professors, graduate students, and PhD-holding IMO medalists, and routinely took their authors weeks to solve independently. Each problem undergoes double-blind verification by two independent domain experts who must solve the problem from scratch, and yields a unique, closed-form solution assessed by programmatic verifiers. We evaluate frontier models as unconstrained research agents, with full access to coding tools, search, and open-ended reasoning, using an unbiased statistical estimator computed over 100 independent runs per problem. Our results reveal that all frontier models currently score below 10\%, exposing a substantial gap between olympiad-level problem solving and genuine research-level mathematical reasoning. By keeping the benchmark fully private, we ensure that measured performance reflects authentic mathematical capability rather than memorization of training data.
\end{abstract}

\section{Introduction}

Five years ago, Surge helped create GSM8K~\citep{cobbe2021gsm8k}, the first mathematical reasoning benchmark for large language models. At the time, the tasks focused on grade-school math and GSM8K became one of the most widely cited benchmarks because it exposed a fundamental gap between fluent language and actual reasoning.

The frontier has since moved dramatically. The year 2025 marked an important moment for AI and mathematics. Google DeepMind's Gemini with Deep Think scored 35 out of 42 points on the 2025 International Mathematical Olympiad (IMO), officially achieving gold-medal standard as certified by IMO coordinators~\citep{deepmind2025imo}. DeepSeekMath-V2 achieved gold-level performance on IMO 2025 and scored 118/120 on Putnam 2024~\citep{deepseekmath2025v2}. These achievements followed a rapid progression: AlphaProof solved the hardest problem at IMO 2024~\citep{alphaproof2025}, and performance on the American Invitational Mathematics Examination (AIME) approached near-perfect accuracy, with o4-mini achieving 99.5\% pass@1 and 100\% consensus@8 on AIME 2025 when given access to a Python interpreter~\citep{openai2025o3o4}. The emergence of reasoning-focused models such as OpenAI's o1~\citep{openai2024o1} and DeepSeek-R1~\citep{deepseekr1}, which apply reinforcement learning to develop extended chains of reasoning, has been a key driver of these gains.


The IMO is deliberately limited to four domains: Algebra, Combinatorics, Geometry, and Number Theory. These foundational areas were specifically chosen because they require minimal advanced machinery; calculus, for instance, is strictly excluded. Because the available tools are limited, IMO problems are designed to reward lateral thinking, often hinging on a single key insight that renders the solution tractable. While IMO problems are incredibly clever, they are fundamentally designed to be solved in a few hours using known tools. The distance between this style of problem solving and the sustained, multi-step theoretical reasoning characteristic of professional mathematical research is vast.

This gap motivates a new evaluation paradigm. While benchmarks such as GSM8K~\citep{cobbe2021gsm8k}, MATH~\citep{hendrycks2021math}, and Omni-MATH~\citep{gao2024omnimath} have progressively raised the difficulty bar, they remain largely confined to competition-style problems. The saturation of existing benchmarks, combined with growing evidence of data contamination in public evaluations~\citep{mirzadeh2024gsmsymbolic,srivastava2024functional,oren2024proving}, underscores the need for private, rigorously constructed benchmarks at the research frontier.

We introduce \bench{}, a benchmark of 25 extreme-tier mathematical problems designed to evaluate AI not on competition puzzles, but on PhD-level research mathematics. We collaborated with Ivy League mathematics professors, graduate students, and PhD-holding IMO medalists to gather problems they encountered in their own research. These problems routinely took their authors weeks to solve independently. Authors noted that their own graduate students and colleagues would struggle to solve these problems on their own.

Our contributions are:

\begin{enumerate}[leftmargin=2em]
\item \textbf{Research-level mathematical benchmark.} We introduce \bench{}, comprising 25 problems spanning diverse mathematical domains including areas that require understanding of variational principles, measure theory, stability analysis, manifolds, and advanced algebraic structures.

\item \textbf{Double-blind, from-scratch verification.} Every problem is independently verified by two domain experts who must solve the problem from scratch before confirming validity.

\item \textbf{Rigorous, unconstrained evaluations.} \bench{} 
evaluates true, unconstrained AI research agents with full access to 
coding tools, search, and open-ended reasoning. We run each frontier 
model 100 times per problem and compute pass rates using the unbiased 
estimator of \citet{chen2021codex}. All models currently score below 
10\%.

\item \textbf{Fully private and uncontaminated.} The dataset is kept strictly private to ensure a fully unbiased evaluation for all frontier labs.
\end{enumerate}

\section{Related Work}
 
The landscape of mathematical reasoning benchmarks has evolved rapidly, tracing a clear difficulty progression from elementary arithmetic to research-level problems.
 
\textbf{Elementary and competition-level benchmarks.} GSM8K~\citep{cobbe2021gsm8k} introduced 8,500 grade-school math word problems requiring 2--8 reasoning steps, alongside the verifier-based evaluation paradigm. The MATH dataset~\citep{hendrycks2021math} raised the bar significantly with 12,500 competition-level problems across seven categories sourced from AMC, AIME, and other competitions. When introduced, the best models achieved roughly 7\% accuracy; frontier models now exceed 90\%, rendering the benchmark effectively saturated. MMLU~\citep{hendrycks2021mmlu} includes mathematics-related subjects among its 57 domains, spanning elementary through college-level abstract algebra.
 
\textbf{Olympiad-level benchmarks.} Several recent benchmarks target olympiad-level difficulty. Omni-MATH~\citep{gao2024omnimath} contains 4,428 problems across 33 sub-domains sourced from competitions including USAMO, APMO, and Putnam. OlympiadBench~\citep{he2024olympiadbench} provides 8,476 bilingual problems in mathematics and physics drawn from international olympiads. OlymMATH~\citep{olympmath2025} targets olympiad-level reasoning across multiple difficulty tiers. JEEBench~\citep{arora2023jeebench} uses 515 problems from India's JEE-Advanced examination. MathOdyssey~\citep{fang2024mathodyssey} contributes expert-crafted problems spanning high school to university level. MathArena~\citep{balunovic2025matharena} evaluates models on recently released competition problems with rigorous contamination controls. The AI Mathematical Olympiad (AIMO) Prize~\citep{aimoprize} has further catalyzed progress by awarding prizes for publicly shared models that solve olympiad-level problems.
 
\textbf{Graduate and research-level benchmarks.} GPQA~\citep{rein2023gpqa} provides 448 expert-crafted multiple-choice questions in physics, chemistry, and biology at a graduate level where domain experts achieve only 65\% accuracy. Humanity's Last Exam~\citep{phan2026hle} crowdsourced 2,500 expert-level questions across dozens of academic disciplines; frontier models scored below 10\% at launch. GHOSTS~\citep{frieder2023ghosts} was among the first benchmarks curated by working mathematicians to target graduate-level mathematics. TheoremQA~\citep{chen2023theoremqa} tests knowledge of over 350 theorems across mathematics, physics, and finance. ARB~\citep{sawada2023arb} targets graduate and expert-level reasoning across multiple domains. SciBench~\citep{wang2023scibench} evaluates college-level scientific problem solving with free-response questions drawn from standard textbooks. MathBench~\citep{liu2024mathbench} spans 3,709 problems across five progressive stages from arithmetic to college mathematics.
 
\textbf{Formal mathematics benchmarks.} MiniF2F~\citep{zheng2022minif2f} contains 488 problems formalized across Lean, Metamath, Isabelle, and HOL Light. ProofNet~\citep{azerbayev2023proofnet} provides 371 parallel examples of formal and natural-language theorem statements from undergraduate textbooks. PutnamBench~\citep{tsoukalas2024putnambench} offers 1,692 hand-constructed formalizations of 640 Putnam competition theorems. DeepSeek-Prover-V2~\citep{deepseekprover2025} recently advanced formal theorem proving by combining reinforcement learning with subgoal decomposition in Lean~4.
 
\textbf{FrontierMath.} FrontierMath~\citep{glazer2024frontiermath}, developed by Epoch AI, contains hundreds of expert-authored problems spanning major areas of modern mathematics. Its public paper reported that then-current state-of-the-art AI models solved under 2\% of problems overall, but newer agentic systems have substantially improved on the benchmark, with Google's AI co-mathematician reporting 48\% on FrontierMath Tier 4~\citep{zheng2026comathematician}.
 
\textbf{AI performance on mathematical olympiads.} AlphaGeometry~\citep{trinh2024alphageometry} solved 25 of 30 historical IMO geometry problems, matching the average gold medalist. AlphaProof~\citep{alphaproof2025}, combined with AlphaGeometry~2~\citep{chervonyi2025alphageometry2}, scored 28/42 at IMO 2024 (silver medal; the gold cutoff was 29). By IMO 2025, Gemini with Deep Think became the first AI system officially certified at gold-medal standard by IMO coordinators~\citep{deepmind2025imo}. DeepSeekMath-V2 achieved gold-level scores on IMO 2025 and CMO 2024, and scored 118/120 on Putnam 2024~\citep{deepseekmath2025v2}. In formal mathematics, Axiom Math reported that AxiomProver solved all 12 problems from the 2025 William Lowell Putnam Mathematical Competition in Lean~4.21.0, with 8 solved by the end of the competition and the remaining 4 solved in the following days~\citep{axiommath2026putnam}.
 
\bench{} complements existing benchmarks in several ways: it is independently constructed, fully private, uses double-blind from-scratch expert verification, and evaluates models as unconstrained research agents.

\section{Benchmark Design}

\subsection{Design Philosophy}

\bench{} differs from competition-style benchmarks in a few important ways. While IMO problems are extremely clever, they are still designed to be solved in a few hours using known tools and mathematical machinery. The problems in \bench{} operate in the domain of PhD-level research, where solving a single problem can take weeks and draws on more specialized theory and mathematical tools.

While every problem has a definite, verifiable answer, the solutions demand long chains of theoretical reasoning, each step building on the last, often pulling from multiple areas of advanced mathematics simultaneously.

\subsection{Problem Construction}

\bench{} comprises 25 problems authored by advanced mathematicians actively engaged in mathematical research. Contributors were asked to draw on problems they encountered in their own research: problems that routinely took them weeks to solve independently. Multiple authors noted that their own graduate students and colleagues would struggle to solve these problems on their own.

Each problem satisfies the following requirements:

\begin{itemize}[leftmargin=2em]
\item \textbf{Unambiguous answer.} Every problem yields a unique, closed-form solution. There is no partial credit and no subjective judgment: the answer is either correct or incorrect.
\item \textbf{Programmatic verification.} When the solution admits multiple equivalent representations, programmatic verifiers assess correctness automatically.
\item \textbf{Research-level difficulty.} Problems require deep domain knowledge and multi-step theoretical reasoning that goes substantially beyond what is testable in competition settings.
\end{itemize}

\subsection{Verification Protocol}

Every problem in \bench{} was subjected to a double-blind verification protocol:

\begin{enumerate}[leftmargin=2em]
\item Two independent domain experts, who were not shown the author's solution in advance, are assigned to verify each problem.
\item Each verifier must solve the problem from scratch and arrive at the correct answer through their own reasoning before confirming a problem's validity.
\item The verifiers also assess problem quality, checking for ambiguity, underspecification, and appropriate difficulty calibration.
\end{enumerate}

This double-blind protocol goes beyond the standard practice of relying on problem authors to self-certify their solutions, providing a stronger guarantee that each problem has a unique correct answer that can be independently derived by multiple experts. Problems that failed verification, due to ambiguity, errors, or insufficient difficulty, were revised or excluded.

\subsection{Privacy}

To ensure a fully unbiased evaluation for all frontier labs, the dataset is kept strictly private. Public benchmarks, however well-intentioned, are vulnerable to leakage and contamination~\citep{xu2024contamination,zhou2023benchmark}. A benchmark that has been seen, even indirectly, is a benchmark that has been compromised. Labs wishing to evaluate their models may submit them through a controlled evaluation service.

\subsection{Unconstrained Agent Evaluation}

Existing benchmarks can force models into rigid, automated evaluation loops. \bench{} evaluates unconstrained AI research agents. Models are given full access to coding tools (Python interpreter), search capabilities, and open-ended reasoning with no artificial constraints on interaction format or token budget. This design reflects our belief that measuring research-level mathematical capability requires allowing models to operate as they would in a genuine research setting.

\section{Illustrative Problem}

To convey the character and difficulty of \bench{}, we present one sample problem. This problem illustrates several key properties of the benchmark: it involves advanced mathematical objects, requires deep familiarity with specialized theory, and demands sustained multi-step reasoning that a domain expert estimated would take 40--50 hours to complete from scratch.

\textbf{Problem overview.} The problem concerns the classification of multibasic $A$-modules over the ring of Hahn series with real-valued valuation and residue field $\mathbb{F}_2$. The field $K$ of Hahn series in indeterminate $t$ with value group $\mathbb{R}$ is considered as a module over its subring $A$ of elements with non-negative valuation. Special $A$-modules, termed \emph{basic} (quotients of submodules of $K$) and \emph{multibasic} (finite direct sums of basic modules), are defined, with the property that every multibasic $A$-module has a unique decomposition into a direct sum of basic submodules. The problem asks for the number of distinct isomorphism classes of multibasic $A$-modules $M$ satisfying three structural conditions involving the endomorphism ring and a dimension function on associated $\mathbb{F}_2$-vector spaces.

\textbf{Discussion.} Hahn series with real-valued valuation are formal infinite series in which the exponents may be any real numbers. A key insight in the solution is that submodules of the Hahn series field behave very simply with respect to the valuation: any submodule is determined by which powers $t^q$ it contains, forcing the submodule to correspond to a cut in $\mathbb{R}$. As a result, every basic module $L/N$ must come from a small list of canonical possibilities. Because multibasic modules decompose uniquely into basic pieces, the classification problem reduces to determining which combinations of these building blocks are allowed.

The three conditions in the problem then act as filters, each eliminating a different family of candidates through a qualitatively distinct algebraic mechanism. The solution draws on a diversity of mathematical ideas: classifying $A$-submodules of $K$ via the valuation, applying the tensor-hom adjunction to determine how tensor products and hom functors interact with multibasic modules, using ring-theoretic properties to constrain which basic summands can appear, and finally reducing to a finite case analysis with a combinatorial count.

\section{Experimental Setup}

\subsection{Models Evaluated}

We evaluated major frontier AI models. All models were evaluated through their respective APIs, with full access to coding tools (Python interpreter), search capabilities, and open-ended reasoning. No artificial constraints were imposed on interaction format or token budget. This setup ensures that measured performance gaps reflect genuine mathematical reasoning limitations rather than implementation bottlenecks.

\subsection{Evaluation Protocol}

For each problem, we ran every model 100 times independently and computed pass rates using the unbiased statistical estimator introduced by \citet{chen2021codex}. Given $n$ total samples and $c$ correct samples, the $\passkk$ estimator is:
\[
\text{pass@}k = 1 - \frac{\binom{n-c}{k}}{\binom{n}{k}}
\]

This estimator provides unbiased measurements of model capability at any sampling budget $k$, computed from a fixed pool of $n = 100$ independent attempts. The resulting difficulty assessments are not based on a small number of attempts or selected runs; they reflect stable, reproducible measurements.

\section{Results}

\subsection{Overall Performance}

\begin{figure}[t]
\centering
\begin{tikzpicture}
\begin{axis}[
    xbar,
    width=12cm,
    height=9cm,
    xlabel={Pass Rate (\%)},
    xlabel style={font=\small},
    symbolic y coords={%
      Claude Opus 4.5,
      GPT 5.2,
      DeepSeek V3.2,
      Kimi K2.5,
      Gemini 3 Pro,
      Claude Opus 4.6,
      Gemini 3.1 Pro},
    ytick=data,
    yticklabel style={font=\small},
    xmin=0, xmax=9,
    xtick={0,1,2,3,4,5,6,7,8},
    xticklabel={\pgfmathprintnumber{\tick}\%},
    bar width=14pt,
    nodes near coords={\pgfmathprintnumber{\pgfplotspointmeta}\%},
    nodes near coords align={horizontal},
    every node near coord/.append style={font=\small, anchor=west},
    enlarge y limits=0.12,
    axis line style={draw=gray!50},
    tick style={draw=gray!50},
    grid=major,
    grid style={gray!15},
    major tick length=0pt,
]
\addplot[
    xbar,
    draw=none,
    fill=gray,
    point meta=explicit,
] table [
    x=val, y=model, meta=val
] {
model                val
{Claude Opus 4.5}    2
{GPT 5.2}            2
{DeepSeek V3.2}      3
{Kimi K2.5}          4
{Gemini 3 Pro}       4
{Claude Opus 4.6}    6
{Gemini 3.1 Pro}     6
};
\end{axis}
\end{tikzpicture}
\caption{$\passone$ across frontier models on \bench{}. All models score below 10\%, confirming that research-level mathematics remains substantially beyond current capabilities.}
\label{fig:results}
\end{figure}
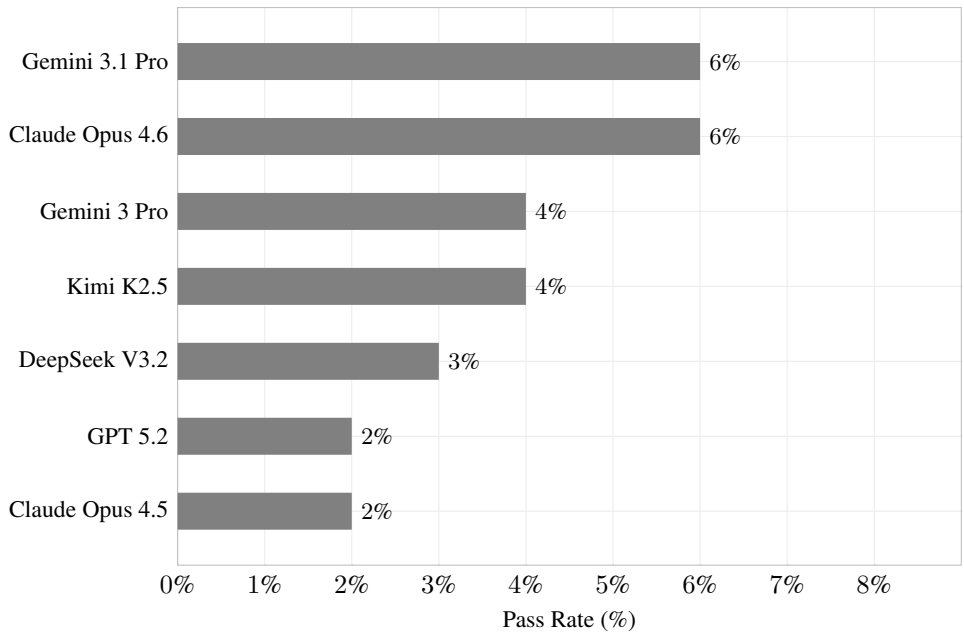

Table~\ref{tab:results} and Figure~\ref{fig:results} present our primary results across all 25 problems.

\begin{table}[t]
\centering
\caption{Frontier model performance on \bench{}. Pass rates are computed from 100 independent runs per problem using the unbiased estimator of \citet{chen2021codex}. All models were evaluated as unconstrained agents with access to coding tools and search.}
\label{tab:results}
\vspace{0.5em}
\begin{tabular}{llc}
\toprule
\textbf{Model} & \textbf{Lab} & \textbf{pass@1 (\%)} \\
\midrule
Gemini 3.1 Pro & Google & 6 \\
Claude Opus 4.6 & Anthropic & 6 \\
Gemini 3 Pro & Google & 4 \\
Kimi K2.5 & Moonshot AI & 4 \\
DeepSeek V3.2 & DeepSeek & 3 \\
GPT 5.2 & OpenAI & 2 \\
Claude Opus 4.5 & Anthropic & 2 \\
\bottomrule
\end{tabular}
\end{table}

The central finding is important: \textbf{all frontier models currently score below 10\% on \bench{}, even with full access to coding tools and search.} This is in sharp contrast to olympiad-level benchmarks, where the same models approach or exceed human gold-medal performance. This gap helps confirm that research-level mathematics, and the kind of long-horizon, multi-step theoretical reasoning that characterizes PhD-level work, remains beyond current model capabilities.

\subsection{Comparison with Competition-Level Performance}

To help contextualize these results, we note the performance of the same model generation on competition-level mathematics. The models evaluated here, or their close variants, achieve near-perfect scores on AIME problems and gold-medal-level performance on IMO problems. The dramatic drop from near-100\% on AIME to below 10\% on \bench{} illustrates the qualitative difference between competition mathematics and research-level problems. Competition problems, however difficult, can often be resolved through a single key insight applied with fairly elementary tools. \bench{} problems require sustained theoretical reasoning over weeks of effort, drawing on specialized knowledge that can extend well beyond the competition canon.

\subsection{Analysis of a Representative Failure Mode}

Beyond aggregate pass rates, qualitative analysis of model failures reveals important patterns in how current systems break down on research-level mathematics. We present a representative failure on the illustrative problem from Section~4 to demonstrate an important class of errors.

\textbf{Model failure.} Rather than working within the $A$-module framework specified by the problem, the model reinterpreted the entire problem in terms of an inapplicable theory of ``generalized scales.'' Specifically, it incorrectly treated conditions (i) and (ii), constraints on the endomorphism ring of $M$, as the definition of a ``basic scale,'' and misinterpreted condition (iii) as specifying the ``support'' of $M$, when in fact the relevant notion of support is intrinsic to the Hahn series construction. To justify its reasoning, the model fabricated a nonexistent classification theorem, attributing it to a fictitious reference (``M.\ Getz, Theorem 4.14 on Generalized Scales''). Applying this fabricated theorem, the model arrived at an answer of $2^{299}$, which is off by orders of magnitude from the correct answer.

\textbf{Broader pattern.} This failure shows a recurring pattern observed across \bench{} evaluations: when confronted with problems requiring specialized theoretical frameworks, models may substitute a superficially related but inapplicable framework and fabricate supporting results to complete the reasoning chain. The model's output reads as structurally coherent: it identifies the problem as a classification task, proposes a theoretical framework, invokes a theorem, and computes a numerical answer, but the entire reasoning chain is built on a misidentified foundation.

\section{Discussion}

\subsection{Why Competition Math Is Not Enough}

The contrast between olympiad-level and research-level performance reveals a fundamental distinction in mathematical reasoning. Problems in \bench{} demand deep familiarity with theory and the ability to reason through and combine multiple advanced results. A model that can identify the key insight in an IMO problem may nonetheless be unable to reason about the Eynard--Orantin topological recursion or classify multibasic modules over Hahn series rings.

This distinction carries important implications for how we interpret benchmark saturation. The saturation of competition-level benchmarks does not imply that mathematical reasoning is solved. It indicates that a particular style of mathematical reasoning, one that rewards lateral thinking with more elementary tools, is within reach of current systems. The broader and deeper domain of research mathematics largely still remains beyond current capabilities.

\subsection{Implications for AI-Assisted Mathematical Research}

The results carry both encouraging and cautionary implications for the prospect of AI systems contributing to mathematical research. The rapid generational improvement observed on competition mathematics suggests that continued scaling and targeted training can yield meaningful progress. On the other hand, performance below 10\% on \bench{} means that even the best models fail on the vast majority of research-level problems, making current systems unreliable as autonomous mathematical reasoning agents.

A more realistic near-term application may be AI-assisted research, in which human mathematicians use AI systems as computational assistants for specific subtasks while verifying outputs against their own expertise. This mirrors the trajectory observed in other domains of AI deployment, where practitioners deliberately constrain agent autonomy to maintain reliability~\citep{pan2025measuring}. Tools such as Lean Copilot~\citep{song2024leancopilot} exemplify this collaborative approach in the context of formal theorem proving.

\subsection{Toward Moonshot Mathematics}

\bench{} problems, however difficult, still have known solutions. The true moonshots of mathematical research require formulating conjectures, building novel frameworks, and navigating spaces in which the existence of an answer is itself unknown. We view \bench{} as a necessary intermediate evaluation along this trajectory. Reliable performance on research-level problems with known solutions is a prerequisite for any system aspiring to contribute to open mathematical research.

\section{Conclusion}

We introduced \bench{}, a private benchmark of 25 extreme-tier mathematical problems for research-level reasoning. Our principal findings are:

\begin{itemize}[leftmargin=2em]
\item All frontier models currently score below 10\% on \bench{}, revealing a vast gap between olympiad-level problem solving and research-level mathematical reasoning.

\item The double-blind, from-scratch expert verification protocol and fully private evaluation design ensure that measured performance reflects genuine mathematical capability rather than memorization.

\item Evaluating models as unconstrained research agents rather than constraining them to rigid evaluation loops, provides a more faithful measure of AI's capacity for open-ended mathematical reasoning.

\item Qualitative analysis of failures reveals that models can substitute inapplicable theoretical frameworks and fabricate supporting results, producing structurally coherent but substantively wrong reasoning chains.
\end{itemize}

Having built the baseline the field relies on with GSM8K, we are now defining its ceiling with \bench{}. AI's success on the IMO marks a beginning, not an end. \bench{} provides a rigorous, contamination-resistant measurement of progress toward the mathematical moonshots that matter.

\section*{Acknowledgments}

We thank the mathematicians and researchers who contributed problems to \bench{} and the domain experts who participated in the double-blind verification protocol. Their expertise and rigor are the foundation of this benchmark.

\bibliographystyle{plainnat}
\bibliography{riemannbench_refs}

\end{document}